\pdfoutput=1

\documentclass[11pt]{article}
\usepackage{booktabs} 
\usepackage{float}

\usepackage{acl}
\usepackage{longtable}
\usepackage{times}
\usepackage{latexsym}
\usepackage{pgffor}
\usepackage{caption} 
\usepackage{subcaption} 

\usepackage{caption}
\usepackage{geometry} 
 
\usepackage{tabularx}
\usepackage{array}

\usepackage[T1]{fontenc}

\usepackage[utf8]{inputenc}

\usepackage{microtype}

\usepackage{inconsolata}

\usepackage{algorithmic}
\usepackage{caption}
\usepackage{graphicx}
\usepackage{textcomp}
\usepackage{xcolor}
\usepackage{tabularx}
\usepackage{booktabs}
\usepackage{makecell}
\usepackage{hyperref}
\usepackage{multirow}
\usepackage[normalem]{ulem}
\useunder{\uline}{\ul}{}
\usepackage{threeparttable}
\usepackage{listings}
\usepackage{amsmath}
\usepackage[ruled,linesnumbered]{algorithm2e}

\usepackage{geometry}
\usepackage{booktabs} 
\usepackage{tabularx, multicol, multirow} 
\usepackage{pdflscape}
\usepackage{xcolor}
\usepackage{cleveref}
\definecolor{lightblue}{RGB}{191,239,255}
\definecolor{lightred}{RGB}{255,192,203}

\title{$R^3$-NL2GQL: A Model Coordination and Knowledge Graph Alignment Approach for NL2GQL}

\author{
\textbf{Yuhang Zhou}\thanks{Email: yuhangzhou22@m.fudan.edu.cn}\(^{1,2}\) \quad
\textbf{Yu He}\(^{1,2}\) \quad
\textbf{Siyu Tian}\(^{1,2}\) \quad
\textbf{Yuchen Ni}\(^{3}\) \quad
\textbf{Zhangyue Yin}\(^{1}\) \\
\textbf{Xiang Liu}\(^{4}\) \quad
\textbf{Chuanjun Ji}\(^{5}\) \quad
\textbf{Sen Liu}\(^{1,2}\) \quad
\textbf{Xipeng Qiu}\(^{1}\) \quad
\textbf{Guangnan Ye}\thanks{Corresponding Author. Email: yegn@fudan.edu.cn}\(^{1,2}\) \quad
\textbf{Hongfeng Chai}\(^{1,2}\) \\
\\
\(^{1}\)School of Computer Science, Fudan University \\
\(^{2}\)Institute of Fintech, Fudan University \\
\(^{3}\)School of Electronics and Information Engineering, Tongji University \\
\(^{4}\)Tandon School of Engineering, New York University \\
\(^{5}\)DataGrand Inc.
}

\begin{document}
\maketitle
\begin{abstract}

While current tasks of converting natural language to SQL (NL2SQL) using Foundation Models have shown impressive achievements, adapting these approaches for converting natural language to Graph Query Language (NL2GQL) encounters hurdles due to the distinct nature of GQL compared to SQL, alongside the diverse forms of GQL. Moving away from traditional rule-based and slot-filling methodologies, we introduce a novel approach, $R^3$-NL2GQL, integrating both small and large Foundation Models for ranking, rewriting, and refining tasks. This method leverages the interpretative strengths of smaller models for initial ranking and rewriting stages, while capitalizing on the superior generalization and query generation prowess of larger models for the final transformation of natural language queries into GQL formats. Addressing the scarcity of datasets in this emerging field, we have developed a bilingual dataset, sourced from graph database manuals and selected open-source Knowledge Graphs (KGs). Our evaluation of this methodology on this dataset demonstrates its promising efficacy and robustness. 

\end{abstract}
\section{Introduction}\label{sec1}

\begin{table}[!h]  
  \renewcommand\arraystretch{1} 
  \captionsetup{justification=centering}  
  \caption{Some keywords of SQL and GQL (using the nGQL language as an example) showcasing the differences between SQL and GQL.}  
  \centering  
  \small 
  \begin{tabular}{>{\centering\arraybackslash}m{1cm} >{\centering\arraybackslash}m{3cm} >{\centering\arraybackslash}m{2.5cm}}  
    \toprule  
    & \textbf{GQL} & \textbf{SQL} \\  
    \midrule  
    C & \makecell{INSERT VERTEX,\\ INSERT EDGE} & INSERT \\ 
    \addlinespace[5pt]
    R & \makecell{DELETE, DROP} & DELETE \\
    \addlinespace[5pt]
    U & \makecell{ALTER, UPDATE,\\ UPSERT} & UPDATE \\  
    D & \makecell{MATCH, LOOKUP,\\ OPTIONAL MATCH, \\GO,FETCH, SHOW,\\ GET, SUBGRAPH, FIND} & SELECT \\  
    Keywords & \makecell{WHERE, LIMIT, \\SKIP, ORDER BY,\\ YIELD, WITH} & \makecell{WHERE, HAVING,\\ ORDER BY, JOIN} \\  
    Expression & \makecell{count(), max(), \\strcasecmp(),\\ timestamp(), properties()} & \makecell{sum(), ceil(), abs(),\\ lower(), data()} \\  
    \bottomrule  
  \end{tabular}  
  \label{table1}
\end{table}

Graph-based data structures are central to diverse areas such as financial risk management, social networking, and healthcare\cite{yu-etal-2022-diversifying-content,zhang2023multimodal}. To manage this data efficiently, graph databases are widely used, offering an effective means to represent and store complex, interconnected information \cite{b1}. Despite their utility, the intricacy of GQL poses a challenge for those not specialized in the field, making it hard to leverage graph databases for data analysis and application development. Meanwhile, although numerous NL2SQL approaches have shown promise~\cite{pourreza2023dinsql} \cite{dong2023c3} \cite{tai2023exploring}, their direct application to NL2GQL is hindered by the differences in focus and syntactic complexity between SQL and GQL, as shown in \autoref{table1}.

\begin{figure}[htbp]
\centerline{\includegraphics[width=\linewidth]{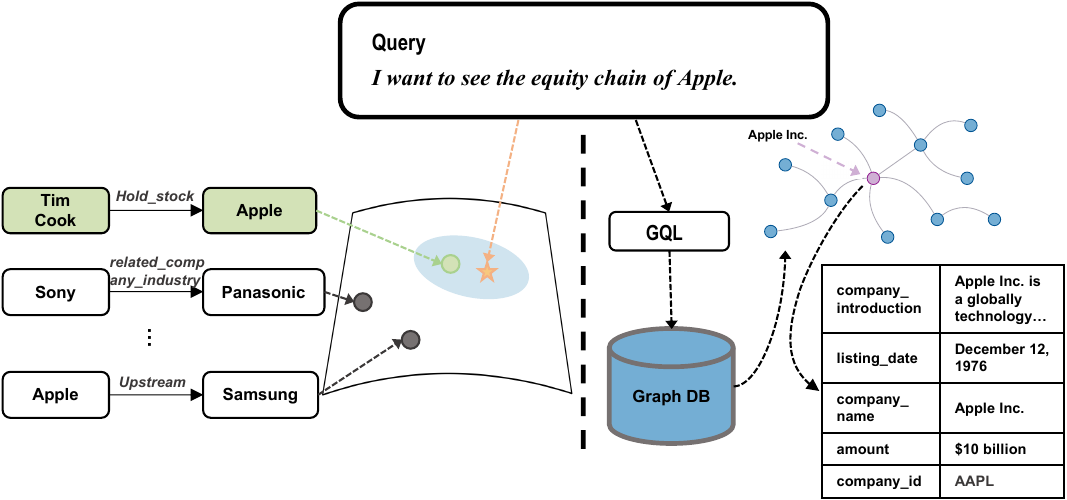}}
\caption{Retrieval algorithm based on triplet vector v.s. GQL-based method.}
\label{fig2}
\end{figure}

Regarding information retrieval in KGs, although triplet vector-based retrieval methods \cite{baek2023direct} offer efficiency and accuracy, they compromise the graph's structural integrity, limiting their utility in complex queries. In contrast, GQL-based methods maintain rich data and logical pathways, bridging the conversational and data-structured worlds, and enhancing the model's interactivity and interpretability, as shown in \autoref{fig2}.

Therefore, implementing a system for the NL2GQL task has become particularly important, but the progress in NL2GQL has been modest, with efforts predominantly concentrating on the Cypher (one type of the GQL). Many solutions, such as Text2Cypher, a Python library, use template-based methods to transform natural language into Cypher, ensuring syntactic correctness but requiring extensive customization for specific data schemas. More recently, SpCQL \cite{b5} introduced the Text to Cypher task and developed the first dedicated dataset, using seq2seq models as a baseline. However, this approach has only achieved a 2\% success rate in generating accurate Cypher queries, indicating significant potential for improvement, while the lack of schemas makes this dataset difficult to apply in real-world environments.

The challenges in NL2GQL stem from several key factors:
\textbf{1) Multiple Model Requirements}: Graph databases complicate GQL formulation with their intricate node-edge structures. Our experiments have shown that a single small model cannot learn GQL syntax through Few-Shot or Finetuning. Larger models, although better at generalizing across schemas, often struggle to align with the specific schemas or data elements within graph databases, leading to errors or hallucinations, making it difficult to solve the NL2GQL task with a single model.
\textbf{2)Limited Resources}: The nascent stage of NL2GQL, contrasted with the well-resourced NL2SQL field, leads to a scarcity of datasets \cite{b2, b3, b4} and tools, hampering research and development efforts in this area.

To address these issues, we developed $R^3$-NL2GQL, combining the specialized insights of fine-tuned smaller models with the broad adaptability of larger ones. The smaller model acts as a ranker and rewriter, while the larger model refines the GQL generation. We also integrated original KG data to optimize alignment, aiming to improve the larger model's zero-shot performance. Facing a lack of NL2GQL datasets, we created a bilingual dataset with thousands of high-quality entries, marking a novel application of Foundation Models in NL2GQL.

We summarize our contributions as follows:

\begin{itemize}


\item \textbf{Model Coordination Approach:} We devised a strategy that harnesses both smaller and larger Foundation Models to overcome NL2GQL obstacles. Our method involves translating schemas into code structures and outlining the basic skeleton for GQL types. In this setup, smaller models function as rankers and rewriters, with a larger model refining the process to enhance GQL generation.

\item \textbf{Bilingual Dataset:} We create a bilingual dataset and set evaluation standards. To the best of our knowledge, this represents the first multi-schema dataset for the NL2GQL task.

\item \textbf{Retrieval and Alignment:} By leveraging node and edge-based representations inherent to database storage mechanics, we address alignment issues between user queries and database schema and elements. Employing a multi-level retrieval mechanism, we connect the relevant data elements to enhance the model’s logical reasoning, thereby improving the accuracy of GQL generation.


\end{itemize}

\begin{figure*}[htbp]
\centerline{\includegraphics[width=0.9\textwidth]{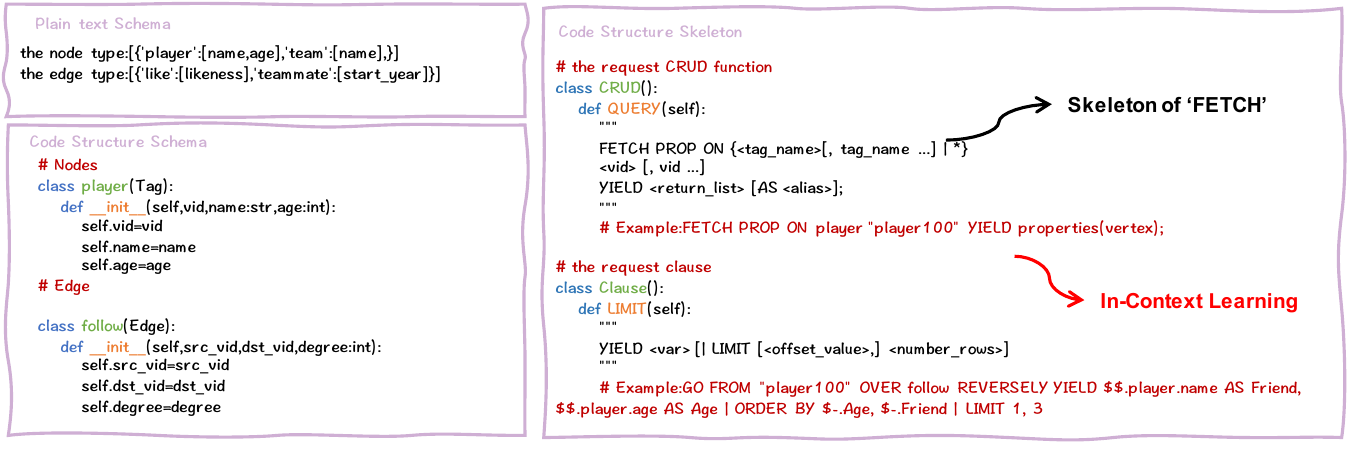}}
\caption{The examples of plain-text schema, code-structure schema, and code-structure skeleton: The plain-text schema serves as the vanilla schema prompt and is written in natural language. The code-structure schema leverages the Python language to re-represent the schema of graphs, with the aim of enhancing the model's inference capabilities. The code-structure skeleton extracts essential keywords and clause information, focusing on GQL.}
\label{fig3}
\end{figure*}

\section{Task Formulation}\label{sec3}

To address the challenge of information loss in natural language schema representations, we devised a novel approach for schema and query formulation in the context.

\subsection{Code-Structured Graph Schema Description}

Transitioning from natural language descriptions to a structured, code-based representation for graph schemas ensures semantic integrity for entities, relationships, and attributes. This involves encapsulating the schema within a Python code structure to reflect the graph’s architecture.

The code structure schema defines various schema structures, consisting of Tag and Edge. Subclasses represent each graph’s schema, utilizing Python features for detailed and precise descriptions: 1) Concept names as Python classes; 2) Class annotations for in-depth explanations; 3) Class inheritance for hierarchical relationships; 4) Init functions for attributes of tags or edges.

The code structured schema, depicted in \autoref{fig3}, enhances the model’s interpretability by maintaining semantic consistency and leveraging the alignment between graph data and object-oriented paradigms \cite{b44}.

\subsection{Code-Structured Skeleton for GQL}
To facilitate the handling of diverse GQL queries, the keywords of GQL are abstracted into a structured framework, aligning them with CRUD operations such as "MATCH" and "FIND" and supplementary clauses such as “LIMIT” and “GROUP.” This framework is also expressed through Python’s class and function constructs, augmented with comments and illustrative examples to demystify the application of each keyword. The design, as shown in right of Figure \hyperlink{fig:3}{\ref{fig3}}, promotes a more clear comprehension and generation of GQLs  by delivering a tangible, example-centric context for every operation within the graph database ecosystem.

\subsection{NL2GQL Task}
A task can be formally represented as:
\begin{equation}
q=f(n, \mathcal{G}, \mathcal{S}),
\end{equation}

\noindent where $\mathcal{G}$ is the data of the given Graph database, including the data format $\mathcal{G}=\left\{\left(s, r, o\right) | s, o \in \mathcal{N}, r \in \mathcal{E} \right\}$, where $\mathcal{N}$ represents node set and $\mathcal{E}$ represents edge set. $\mathcal{S}$ represents the schema of the graph database, $n$ represents the natural language requirements input by the user, and can be segmented according to the token $n=\left\{n_1, n_2, n_3,... ,n_i\right\}$, $q$ represents the final generated GQL.


\section{$R^3$-NL2GQL Framework}\label{sec4}

\begin{figure*}[htbp]
\centerline{\includegraphics[width=0.9\textwidth]{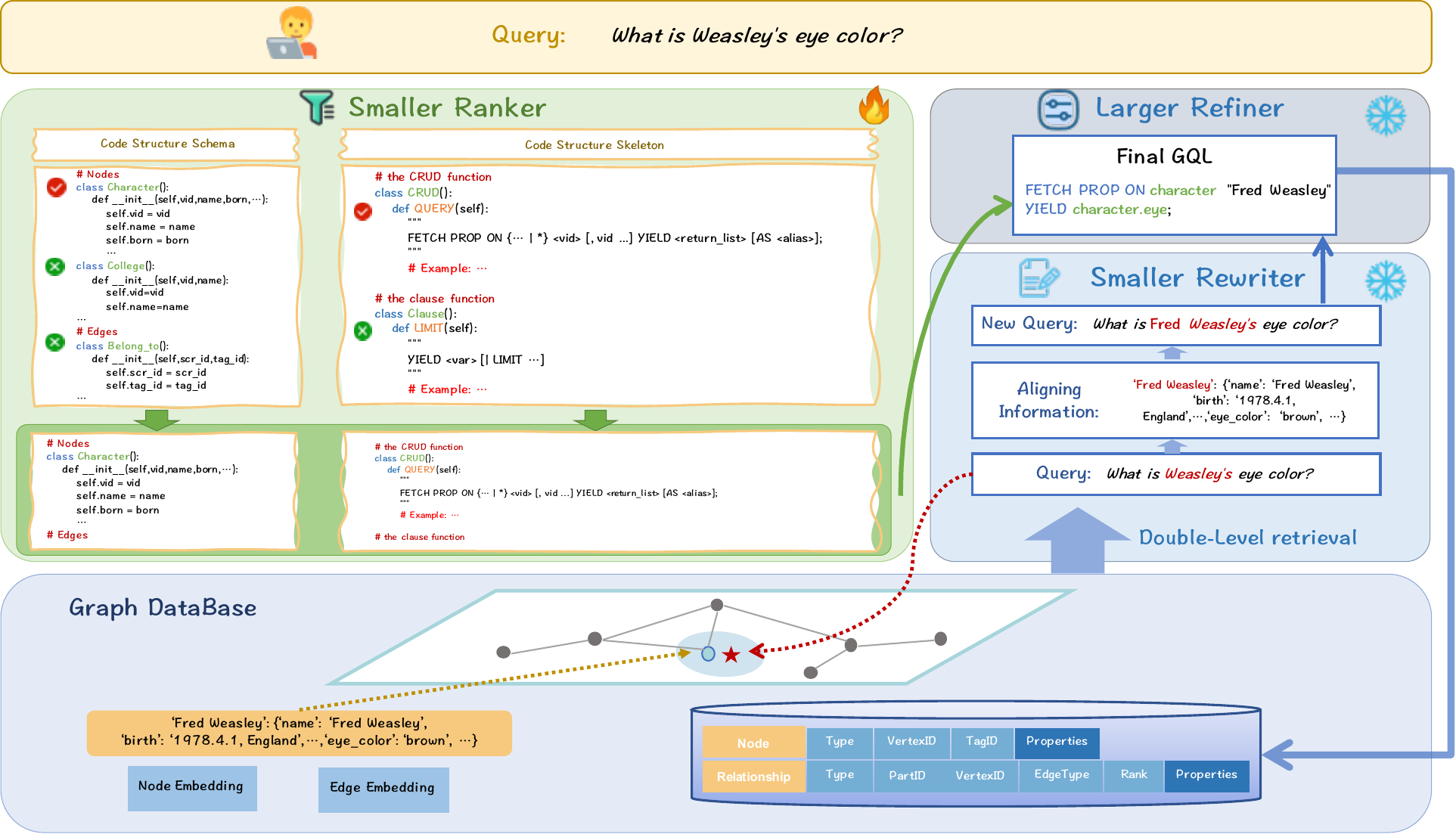}}
\caption{An Overview of $R^3$-NL2GQL: Employing a smaller white-box model as a ranker, it selects required CRUD functions, clauses, and schema from the input. Another smaller white-box model serves as a rewriter, aligning the query with the intrinsic database k-v storage to mitigate the hallucinations. Lastly, a larger model is harnessed for the purpose of generating GQL, capitalizing on its ability in generalization and generation.}
\label{fig4}
\end{figure*}

The $R^3$-NL2GQL framework pioneers a coordination strategy, merging several models to mitigate the limitations of relying on a single model, as illustrated in Figure \hyperlink{fig:4}{\ref{fig4}}. The process initializes with a finely tuned smaller model serving as a ranker, excel at identifying key components like CRUD operations, clauses, and schema classes from the input. To tackle the alignment challenge, another smaller model leverages Few-Shot learning to fetch and validate information against the graph database, functioning as a rewriter to guarantee data precision. The outputs of these models are then further honed by a larger model, tapping into its sophisticated generalization and synthesis capabilities to ultimately generate accurate GQLs.

\subsection{Smaller Foundation Model as Ranker}

The transformation from natural language queries to GQL involves distinct phases, each presenting unique challenges:

\begin{itemize}
    \item \textbf{CRUD Keyword Selection:} Identifying the correct CRUD keywords is foundational, setting the stage for the query structure.
    \item \textbf{Clause Determination:} Following CRUD keyword selection, the next step involves choosing the necessary clauses to construct a coherent query, considering filters, sorting, and other elements aligned with user intent.
    \item \textbf{Node and Edge Identification:} The final phase entails pinpointing the specific nodes and edges to interact with within the GQL schema, ensuring the query fetches the intended data.
\end{itemize}

To address these steps efficiently, we introduce a smaller foundation model as a ranker. Drawing on the benefits of code pre-training, which is considered by some studies to enhance a model's reasoning capabilities\cite{yang2024llm}, we utilize code-structured schemas and skeletons to assist the ranker in its task:

\begin{equation}
    \small
    \text{SCH}_{\text{sub}}, \text{SKE}_{\text{CRUD\&clause}} = \text{ranker}(\text{SCH}, \text{SKE}, n)
\end{equation}

Here, "SCH" and "SKE" represent the code-structured schema and skeleton, while "n" is the natural language query. The output includes a schema subset ($\text{SCH}_{\text{sub}}$) and the necessary keywords and clauses ($\text{SKE}_{\text{CRUD\&clause}}$), both in code structure, ensuring alignment with the query's intent.

A specialized dataset, detailed in Section \ref{sec5}, was developed for training and evaluating the ranker, ensuring its effectiveness in facilitating the NL2GQL conversion process.

\subsection{Smaller Foundation Model as Rewriter}
To guarantee the accurate linkage of corresponding nodes, edges, and schema within the graph data by the generated GQL, we employ a smaller model to serve as the rewriter for precise alignment.

\subsubsection{Aligning Data in Graph Databases}
\begin{figure}[htbp]
\centerline{\includegraphics[width=0.9\linewidth]{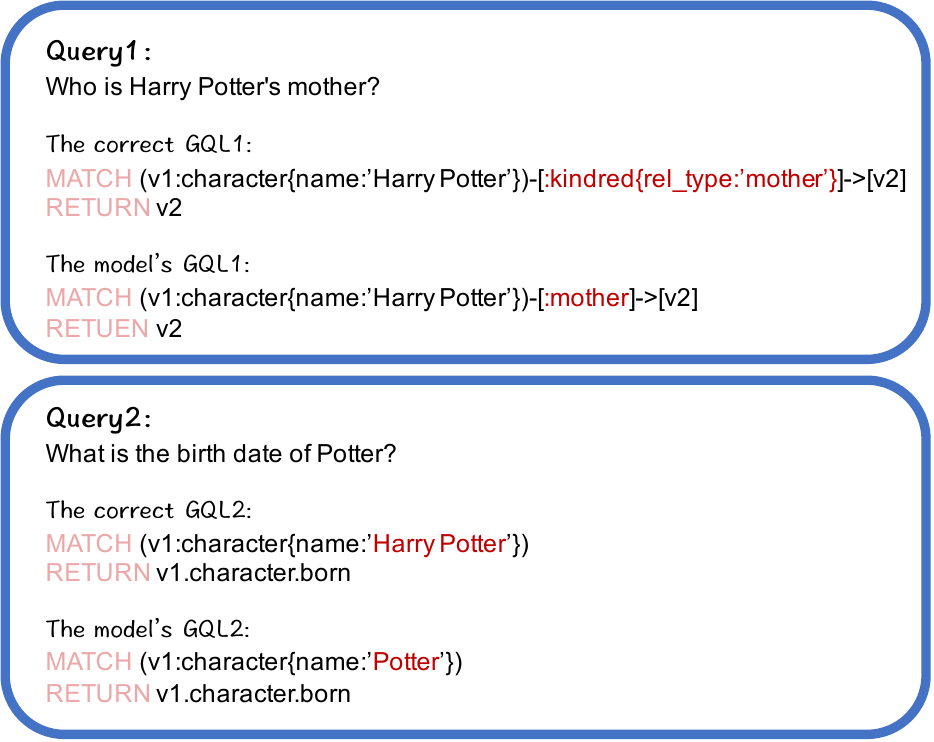}}
\caption{The challenge of aligning user queries with the actual graph data: the error has been marked in red.}
\label{fig5}

\end{figure}

Figure \hyperlink{fig:5}{\ref{fig5}} illustrates the challenge of aligning user queries with the actual graph data, such as mismatches between queried entities and their representations in the database. For example, a query about `Harry Potter's mother' may not directly correspond to the existing graph structure, necessitating adjustments to fit the schema. At the same time, the model may also create node or edge types that are not included in the schema, and this hallucination phenomenon will lead to errors.

\begin{figure*}[h]
\centerline{\includegraphics[width=0.8\textwidth]{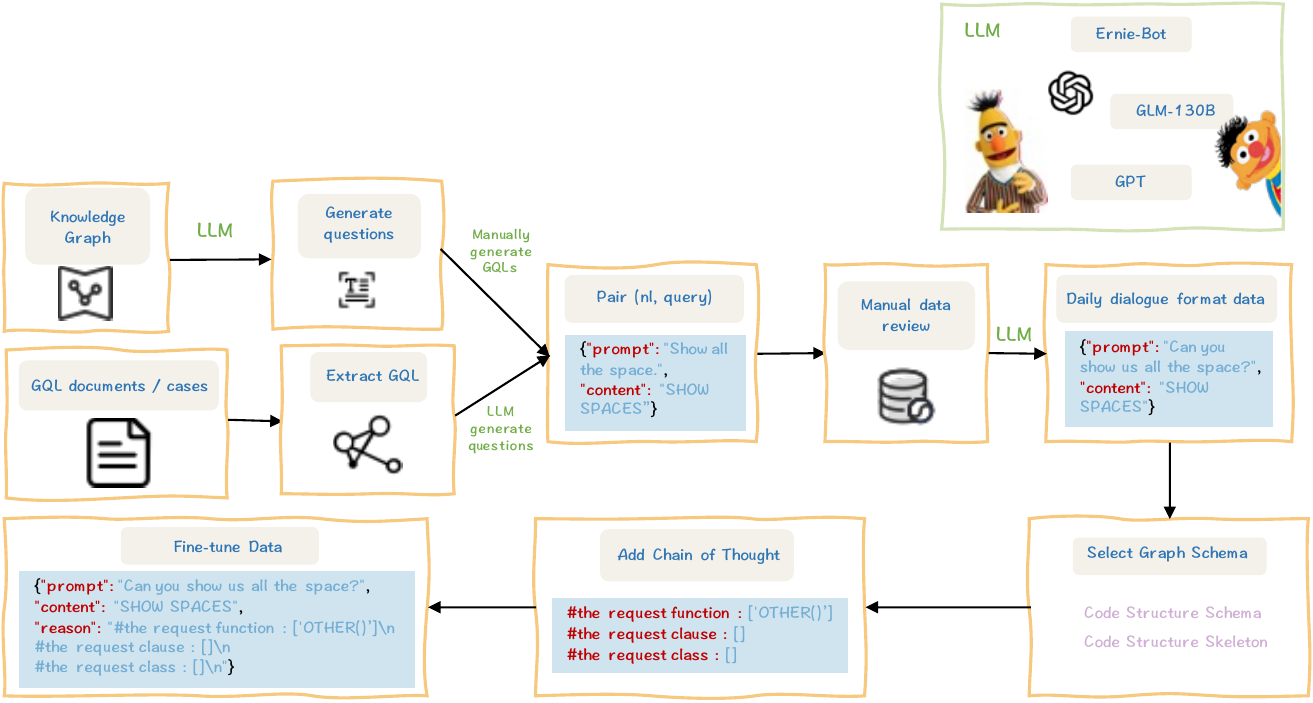}}
\caption{Data construction pipeline}
\label{fig7}
\end{figure*}

\subsubsection{Graph Database Storage Principles}
Graph databases, such as Neo4j, NebulaGraph, and JanusGraph, store data as nodes and edges using distinct storage engines. These systems organize graph data into array-like files, translating them into a ``{node: attributes}, {edge: attributes}'' format, as shown in \autoref{app:GDB}. This storage method aligns with our retrieval methods, minimizing continuous query requests and reducing memory usage during the alignment process.

\subsubsection{Data Retrieval}
The goal of data retrieval is to accurately match the user's query with the corresponding data in the DB, addressing alignment issues. This involves a two-level retrieval and alignment process:

\textbf{Character-Level Alignment}: Utilizing Levenshtein Distance\cite{yujian2007normalized} (Minimum Edit Distance) to calculate the similarity between the query and database entities, defined as \autoref{equ:character}.

\begin{equation}
\label{equ:character}
    U_1 = \frac{\min[\text{len}(Q), \text{len}(I)]}{\text{Levenshtein}(Q, I)}
\end{equation}


where "Q" is the user's input NL query, and "I" represents the data within the graph.

\textbf{Semantic Vector-Based Alignment}: Embedding both the user query and graph data in a dense vector space to facilitate deeper semantic matching, defined as \autoref{equ:semantic}.

\begin{equation}
\label{equ:semantic}
U_2 = \frac{\text{Emb}(Q) \cdot \text{Emb}(I)}{\|\text{Emb}(Q)\| \|\text{Emb}(I)\|}
\end{equation}

This step focuses on rectifying discrepancies between the query and the actual graph data, ensuring the query's alignment with the database's structure.





\subsection{Larger Foundation Model as Refiner}

Positioned as the culminating element in our methodology, the larger model integrates inputs from the preceding smaller models, enhancing GQLs generation. It consolidates code-structured schemas and skeletons identified by the ranker, along with the rewriter's adjusted queries and pertinent retrieval outcomes. This amalgamation, enriched by the larger model's advanced Zero-Shot capabilities, facilitates the creation of refined GQL queries. This synergy between the models amplifies the system's ability to interpret and respond to complex queries with heightened accuracy.

\section{Data Design} \label{sec5}

In contrast to the numerous open-source datasets for NL2SQL tasks, such as Spider and KaggleDBQA \cite{b17}, GQL is deficient in large-scale, diverse-schema datasets that meet real-world industrial requirements. Most existing datasets predominantly focus on Cypher, making it challenging to create a dataset for GQLs.

To address this gap, we developed a multi-schema dataset for NL2GQL. Leveraging Foundation Models' proficiency in generating Cypher, we choose nGQL for our research to evaluate our approach. This section outlines our methodology for defining GQL generation tasks and synthetic data generation, as shown in Figure \hyperlink{fig:7}{\ref{fig7}}.

\subsection{Pair Design}


In constructing the dataset, we avoided directly extracting NL-GQL pairs from GQL documents due to their inability to capture complex human-database interactions. Instead, we used two methods. 1) We manually crafted sample pairs, prioritizing code interpretability over generation, and employed a GQL2NL strategy, using Foundation Models to generate multiple natural language interpretations for each GQL query, followed by manual refinement to closely mimic real-world queries. 2) To include diverse graph schemas, we adapted open-source graph datasets, using their schema and entity information to generate KBQA-style questions with Foundation Models, and then meticulously annotated the GQLs manually to create accurate pairs. These methods resulted in a high-fidelity dataset with numerous NL-GQL pairs, as shown in \autoref{gongshi4}.

\begin{equation}
\label{gongshi4}
    D=Pair(NL_i, GQL_i).
\end{equation}

\subsection{Data Refinement}

The initial dataset may contain inaccuracies and lack linguistic variety, necessitating a phase of data filtering and restructuring. Significant human and computational efforts correct any NL or GQL discrepancies. To enhance naturalness and diversity, we expanded and refined the data. For example, "Find node a" was rephrased to "Hello, I want to find node a, could you assist me by returning its information?" This approach, applied across languages, resulted in a polished and versatile foundational dataset.





\subsection{Incorporating Schema, Skeleton, and Reasoning}

To train the ranker model, we supplemented the training dataset with relevant data. We propose a refined tripartite reasoning framework for GQL formulation, which includes: 1) selecting suitable CRUD operations based on user-input natural language queries, 2) choosing appropriate conditional clauses like LIMIT and WHERE to meet result constraints, and 3) identifying specific node or edge types from the schema for precise GQL construction. This approach results in the final training dataset, as shown in \autoref{gongshi10}, with 'SCH' for 'SCHEMA,' 'SKE' for 'SKELETON,' and 'REA' for 'REASONING'.

\begin{equation}
\small
    D_{train}=\{NL_i, GQL_i, SCH_i, SKE_i, REA_i\}.
\label{gongshi10}
\end{equation}

\subsection{Data Setting}

Through a structured data engineering approach, we constructed a diverse dataset encompassing nine different sectors such as finance, healthcare, sports, and literature, selecting samples from various schemas to enhance the model's generalization capabilities. In each category, we employed the K-Center Greedy \cite{kleindessner2019fair} method to identify the most diverse samples. This approach maintained the original schema distribution, ultimately generating a bilingual dataset of 5000 samples, which was split into training and testing sets at a 4:1 ratio. The test set included schema types absent from the training set to evaluate the model's generalization capabilities.

\section{Experiment}\label{sec6}

 We introduced a multi-tiered evaluation system for NL2GQL tasks, covering aspects from syntax to semantics, detail in \autoref{app:metrics}. Utilizing the dataset, we test the performance of our framework against GPT family counterparts.

\subsection{Settings}\label{sec6.C}


In the absence of established NL2GQL models, we benchmarked against three prominent Foundation Models: text-davinci-003, gpt-3.5-turbo-0613, and GPT-4. These models, extensively trained on diverse textual and code data, served as our baseline using a Vanilla Prompt of natural language-GQL pairs with serialized text schemas. Experiments were conducted in Zero-Shot, One-Shot, and Few-Shot settings, with the latter two involving random selection of examples from training data.

\begin{table*}[h]
    \centering
    \caption{Comparison of the four metrics (\%) among $R^3$-NL2GQL and the GPT family models. The bold numbers denote the best results and the underlined ones are the second-best performance.}
    \label{tab2}
    \begin{tabular}{lcccc}
        
        \toprule
       \textbf{Model} & \textbf{Syntax} & \textbf{Comprehension} & \textbf{Execution} & \textbf{Intra Execution} \\
         & \textbf{Accuracy} & \textbf{Accuracy} & \textbf{Accuracy} & \textbf{Accuracy} \\

        \midrule
        \multicolumn{5}{l}{\textbf{Zero-Shot}} \\
        \quad Vanilla Prompt (text-davinci-003) & 8.59 & 88.17 & 5.44 & 63.28 \\
        \quad Vanilla Prompt (GPT-3.5-turbo-0613) & 6.42 & 88.35 & 4.39 & 68.36 \\
        \quad Vanilla Prompt (GPT-4) & 13.77 & 89.72 & 9.83 & 71.83 \\
        \multicolumn{5}{l}{\textbf{One-Shot}} \\
        \quad Vanilla Prompt (text-davinci-003) & 18.67 & 89.53 & 12.45 & 66.71 \\
        \quad Vanilla Prompt (GPT-3.5-turbo-0613) & 20.45 & 89.15 & 14.45 & 70.65 \\
        \quad Vanilla Prompt (GPT-4) & 25.33 & 90.32 & 19.39 & 76.53 \\
        \multicolumn{5}{l}{\textbf{Few-Shot}} \\
        \quad Vanilla Prompt (text-davinci-003) & 41.16 & 90.01 & 29.79 & 72.37 \\
        \quad Vanilla Prompt (GPT-3.5-turbo-0613) & 28.70 & 90.67 & 21.56 & 75.12 \\
        \quad Vanilla Prompt (GPT-4) & \underline{48.23} & \underline{91.13} & \underline{42.08} & \underline{87.25} \\
        \multicolumn{5}{l}{\textbf{Our}} \\
        \quad $R^3$-NL2GQL (GPT-3.5-turbo-0613) & 36.82 & 90.15 & 30.53 & 82.92 \\
        \quad $R^3$-NL2GQL (GPT-4) & \textbf{57.04} & \textbf{91.57} & \textbf{51.09} & \textbf{89.56} \\
        \bottomrule
    \end{tabular}
\end{table*}

We also evaluated four smaller Foundation Models as ranker and rewriter: LLaMA3-7B\cite{touvron2023llama}, InternLM \cite{team2023internlm}, ChatGLM2 \cite{zeng2022glm}, Flan-T5 \cite{chung2024scaling}, and BLOOM \cite{le2023bloom}, each significantly smaller than GPT family models. To address sampling variability, experiments were repeated thrice for each model, and results were averaged. For the larger Foundation Models, we used OpenAI's API with specific settings (temperature 0.2, top\_p 0.7) to generate nGQLs. The BGE model facilitated embedding during retrieval, with experiments conducted on an NVIDIA A800 GPU using Pytorch 2.0 and Deepspeed. The ranker model was fine-tuned using LoRA (lora\_rank of 8) and optimized with the AdamW optimizer.



\subsection{Main Results}

Table \ref{tab2} showcases the comparative performance between our $R^3$-NL2GQL framework and leading GPT series models across Zero-Shot, One-Shot, and Few-Shot scenarios. Our results indicate that our proposed approach with Zero-Shot excels in the Vanilla Few-Shot setting, underscoring that its performance is not solely reliant on the inherent capabilities of the GPT series models but rather on the reasoning and enhancements integrated into this method. Further examination of the CA metric and outputs from the validation dataset indicates that models with larger parameters demonstrate better understanding and adaptability, particularly in handling intricate schema environments. By harnessing the capabilities of larger models and integrating insights from smaller models, our approach enhances entity linking and generalization, leading to improved performance.

\subsection{Ablation experiment}

We conducted ablation studies to evaluate the contributions of various components within the $R^3$-NL2GQL framework, focusing on the impact of different inputs on the large model's final output. These findings, detailed in \autoref{fig：Ablation}, explored the role of code-structured skeletons as syntax-constrained context prompts, effectively transitioning the Few-Shot methodology to a Zero-Shot paradigm. For a comprehensive analysis, we also included the second-best Few-Shot performance with a Vanilla Prompt (GPT-4) from Table \ref{tab2}. Our proposed Code Prompt showed improvements over the Vanilla Prompt's Few-Shot format across all four metrics, with a 6\% increase in performance on SA and EA.

\begin{figure}[htbp]
\centerline{\includegraphics[width=1\linewidth]{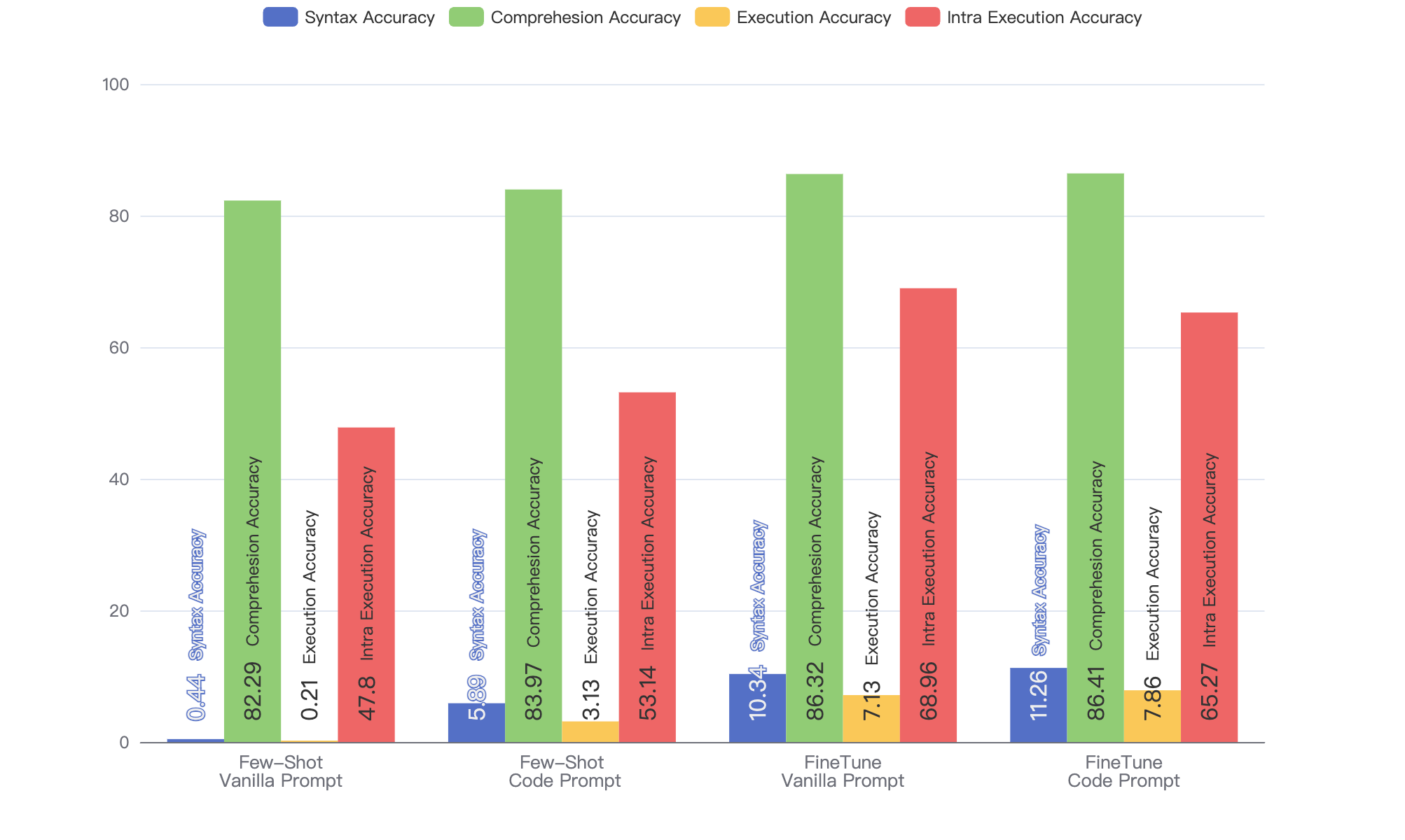}}
\caption{Ablation experiments on smaller models such as LLaMA3-7B, InternLM, ChatGLM2, Flan-T5, and BLOOM.}
\label{fintune}
\end{figure}

\begin{figure*}[h]
\centerline{\includegraphics[width=0.9\textwidth]{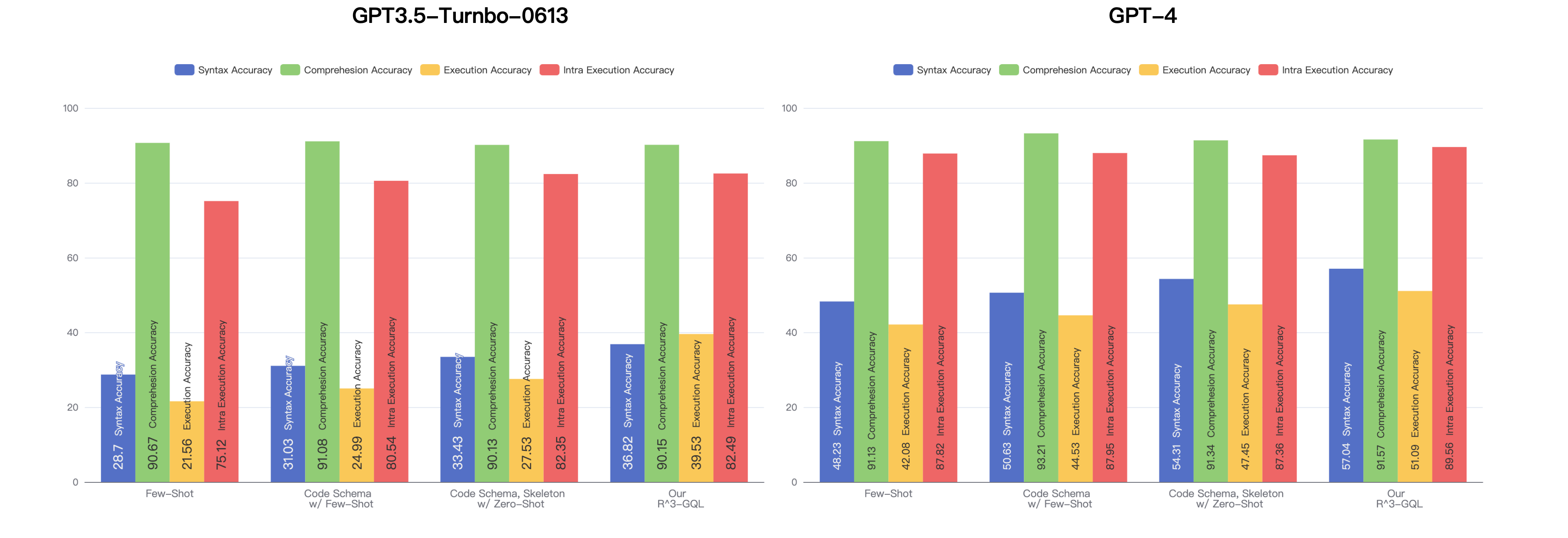}}
\caption{The ablation experiment of GPT-4 and GPT3.5, focus on designing the ablation of each key component.}
\label{fig：Ablation}
\end{figure*}

The results underscored the significant enhancement brought about by incorporating a code-structured schema and skeleton prompt across all models. Replacing the Few-Shot approach with a code-structured skeleton not only refined grammatical accuracy but also enriched the models with a broader spectrum of GQL keywords, diversifying the models' output styles and altering the GQL generation style closer to the standard GQL format. Simultaneously, to validate the capabilities of smaller models, we conducted Few-Shot and fine-tuning experiments on these models, as shown in \autoref{fintune}. The results revealed extremely low SA for these methods. Even after fine-tuning, the SA was only about 10\%, and the IEA metric was below 70\%. This indicates the low generalization and GQL syntax learning abilities of these smaller models, affirming the necessity of collaboration between large and small models.Ultimately, the synergistic use of both larger and smaller models within our framework proved most effective, adeptly synthesizing crucial information and reducing hallucinations to deliver superior results.

\section{Discussions}\label{sec7}

\subsection{Error Analysis}

\begin{figure}[htbp]
\centerline{\includegraphics[width=1\linewidth]{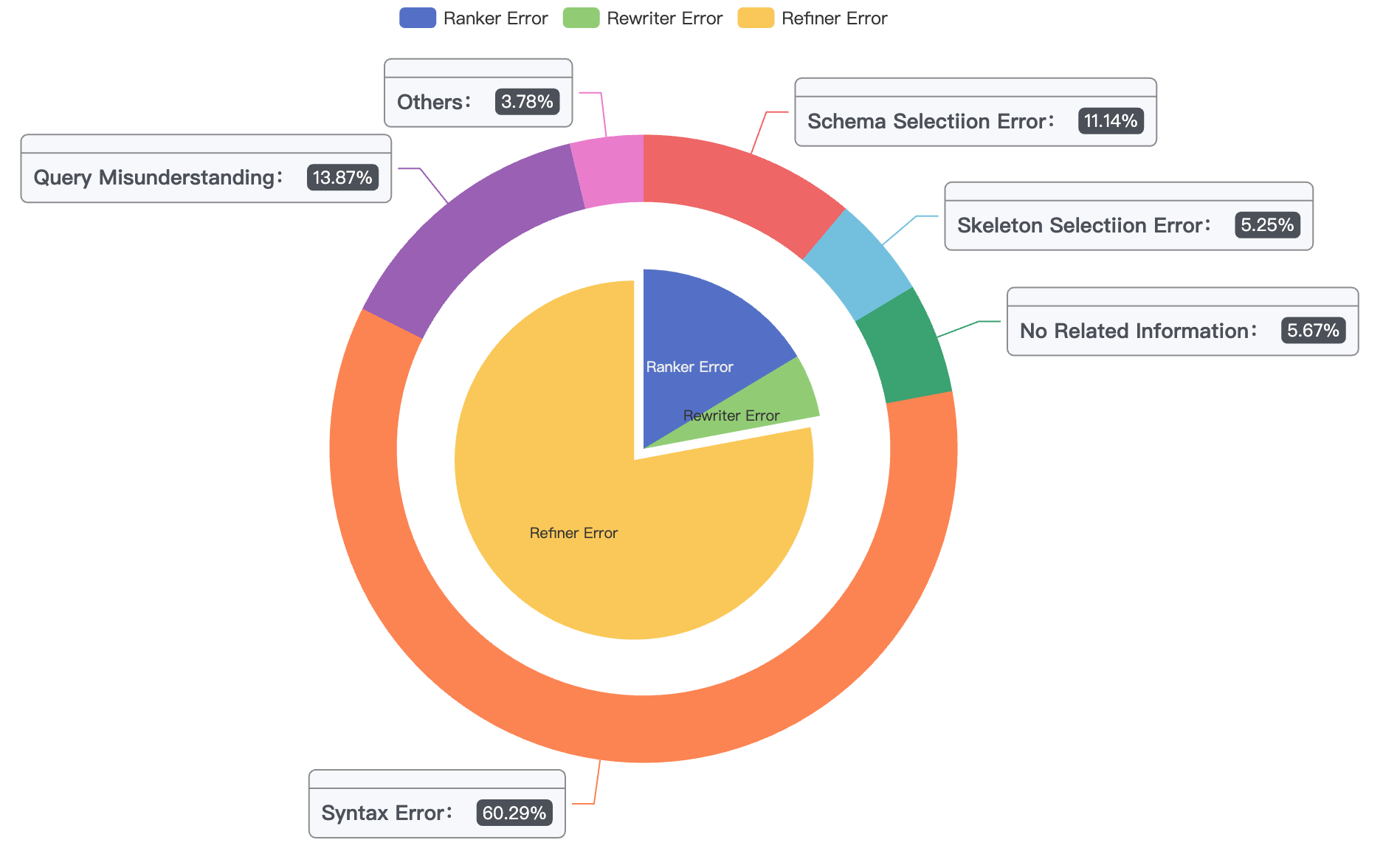}}
\caption{Error Statistical Analysis.}
\label{error_pie}
\end{figure}

Based on Table \ref{tab2} , the EA indicator for $R^3$-NL2GQL is 51.09\%, while the IEA indicators for almost all methods have reached levels above 70\%, with $R^3$-NL2GQL nearly reaching 90\%. This indicates that the vast majority of errors are caused by syntax errors in the generated GQL. We categorize the error types into three major categories and six minor categories, with specific details and examples provided in \autoref{app:error}. Figure \ref{error_pie} presents a statistical analysis of the error information, showing that the majority of errors are caused by Larger Refiner, and in-context learning style struggles to incorporate new GQL syntax into the Foundation Model. Additionally, 13.87\% of errors are caused by misunderstandings of the query. Among the errors in the Ranker, schema selection errors are more likely to affect the final outcome, while the Rewriter demonstrates better performance.

\subsection{Optimal Schema and Skeleton Format for GQL Generation}


The format in which language types, such as code or natural language, are presented plays a pivotal role in a model's ability to grasp the NL2GQL task and comprehend the underlying graph schema. This, in turn, affects its capability to apply these insights to new, unseen scenarios or schemas. Unlike the ambiguous nature of natural language, code language, with its structured syntax and clear execution paradigms, offers a more precise medium for representing instructions and programming constructs. This structured approach, especially in object-oriented languages with features like class inheritance and method definitions, aligns well with graph schema representation, enhancing a model's reasoning capacity for complex tasks, as suggested by recent studies \cite{b44}.

\section{Conclusion}\label{sec9}

Our study presents a novel model coordination framework designed for the NL2GQL task, leveraging the complementary strengths of larger and smaller Foundation Models. By delineating clear roles for each model, we markedly improve the NL2GQL conversion. Additionally, the development of a GQL-specific bilingual dataset underscores the superior performance of our framework. These results pave the way for future advancements in the field of NL2GQL, offering a robust foundation for further exploration and development.

\section*{Limitation and Ethics Statement}
Our study centers on the nGQL query syntax. While analogous languages exist, we have not extended our experimentation to include them. 
Furthermore, the absence of prior assessment standards for NL2GQL tasks means the evaluation criteria we have devised might not be exhaustive.

The dataset used in the paper does not contain any private information. All annotators have received enough labor fees corresponding to their amount of annotated instances. 


\bibliographystyle{acl_natbib}
\bibliography{sample,main}

\clearpage
\onecolumn

\appendix

\section{Difference Between SQL and GQL}
\label{app:GQL&SQL}
Structured Query Language (SQL) and Graph Query Language (GQL) are fundamentally different in their approach to data querying, SQL being tailored for relational databases with its tabular data structure and GQL designed for graph databases which utilize nodes, edges, and properties. SQL provides a declarative approach for users to specify desired data, allowing for complex multi-table join operations and fine-grained control over data retrieval. In contrast, GQL is intuitive for expressing complex relationships and patterns, enabling users to specify the depth and breadth of queries while retrieving granular data, making it particularly suitable for applications with highly interconnected data.

\section{Details of GQL Skeleton}
\label{app:GQLSKE}

GQL incorporates a set of essential keywords within its skeleton, which can be categorized into CRUD operations and clauses. The CRUD operations, such as INSERT, MATCH, UPDATE, and DELETE, facilitate the creation, retrieval, modification, and deletion of data within a graph database. These operations enable users to interact with the database by specifying actions to be performed on the nodes and edges. On the other hand, the clauses in GQL, such as LIMIT, GROUP BY, and WHERE, provide a means to refine and constrain the query results. These clauses allow users to specify conditions, control the number of results returned, and group the data based on certain attributes. The combination of CRUD operations and clauses in GQL empowers users to effectively manipulate and retrieve data from graph databases, catering to a wide range of querying needs.

\setlength{\extrarowheight}{5pt} 
\begin{longtable}{p{0.1\linewidth} p{0.3\linewidth} p{0.5\linewidth}}
    \caption{Some CRUD Keywords in GQL Skeleton}  \\
    \toprule
    \textbf{Keyword} & \textbf{Keyword Meaning} & \textbf{Keyword Example} \\
    \midrule
    \endhead
    CREATE SPACE & Create a new graph database space & CREATE SPACE my\_graph(space\_id: int, ...); \\
    CREATE TAG & Create a vertex label, defining vertex properties & CREATE TAG person(name: string, age: int); \\
    CREATE EDGE & Create an edge type, defining edge properties & CREATE EDGE knows(since: int); \\
    INSERT & Insert new vertices or edges into the database & INSERT VERTEX person(name, age) VALUES "alice":("Alice", 30); \\
    GO & Traverse the database based on specified conditions & GO FROM "alice" OVER knows YIELD \$\$.person.name; \\
    FETCH & Retrieve properties of vertices or edges & FETCH PROP ON person "alice" YIELD person.name, person.age; \\
    LOOKUP & Index-based query operation & LOOKUP ON person WHERE person.age > 25 YIELD person.name; \\
    MATCH & Match graph patterns, used for complex queries & MATCH (p:person)-[:knows]->(f:person) RETURN p.person.name, f.person.name; \\
    UPDATE & Update properties of vertices or edges in the database & UPDATE VERTEX "alice" SET person.age = 31; \\
    UPSERT & Insert or update operation; insert if it does not exist & UPSERT VERTEX "bob" SET person.name = "Bob", person.age = 28; \\
    DELETE & Delete vertices or edges from the database & DELETE VERTEX "bob"; \\
    GET SUBGRAPH & Obtain a subgraph of the graph & GET SUBGRAPH 2 STEPS FROM "alice" YIELD VERTICES AS friends, EDGES AS relationships; \\
    FIND PATH & Find a path between two vertices & FIND SHORTEST PATH FROM "alice" TO "bob" OVER * YIELD path as p; \\
    \bottomrule
\end{longtable}

\setlength{\extrarowheight}{7pt} 
\begin{longtable}{p{0.1\linewidth} p{0.3\linewidth} p{0.5\linewidth}}
    \caption{Some Clauses Keywords in GQL Skeleton}  \\
    \toprule
    \textbf{Keyword} & \textbf{Keyword Meaning} & \textbf{Keyword Example} \\
    \midrule
    \endhead
    GROUP BY & Group results by a variable and apply aggregation functions & GO FROM "player100" OVER follow BIDIRECT YIELD \$\$.player.name as Name | GROUP BY \$-.Name YIELD \$-.Name as Player, count(*) AS Name\_Count \\
    LIMIT & Limit the number of rows returned by a query & GO FROM "player100" OVER follow REVERSELY YIELD \$\$.player.name AS Friend, \$\$.player.age AS Age | ORDER BY \$-.Age, \$-.Friend | LIMIT 1, 3 \\
    SKIP & Skip a number of rows before starting to return rows from a query & MATCH (v:player{name:"Tim Duncan"}) --> (v2) RETURN v2.player.name AS Name, v2.player.age AS Age ORDER BY Age DESC SKIP 1 \\
    SAMPLE & Sample a specified list of steps in a traversal & GO 3 STEPS FROM "player100" OVER * YIELD properties(\$\$).name AS NAME, properties(\$\$).age AS Age SAMPLE [1,2,3] \\
    ORDER BY & Sort the results of a query by one or more expressions & FETCH PROP ON player "player100", "player101", "player102", "player103" YIELD player.age AS age, player.name AS name | ORDER BY \$-.age ASC, \$-.name DESC \\
    WHERE & Filter the results of a query based on specified conditions & MATCH (v:player) WHERE v.player.name == "Tim Duncan" XOR (v.player.age < 30 AND v.player.name == "Yao Ming") OR NOT (v.player.name == "Yao Ming" OR v.player.name == "Tim Duncan") RETURN v.player.name, v.player.age \\
    WITH & Use the results of a match expression for further processing & MATCH p=(v:player{name:"Tim Duncan"})--() WITH nodes(p) AS n UNWIND n AS n1 RETURN DISTINCT n1 \\
    UNWIND & Expand a list and return each element as a separate row & UNWIND [1,2,3] AS n RETURN n \\
    \bottomrule
\end{longtable}

\section{Core Storage of Graph Databases}
\label{app:GDB}
\begin{figure}[htbp]
\centerline{\includegraphics[width=0.9\linewidth]{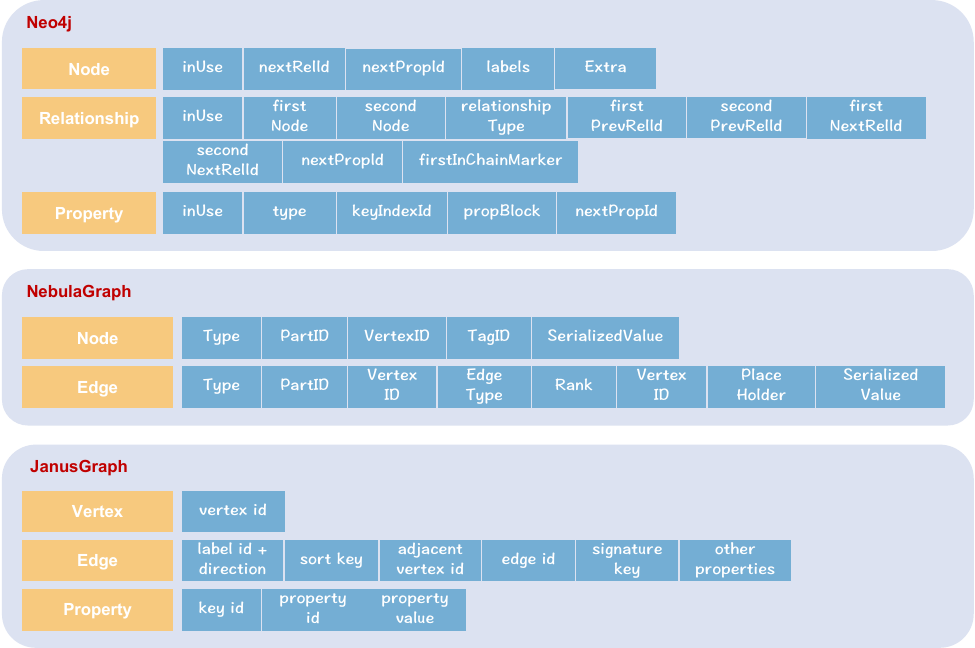}}
\caption{The storage formats of the three graph databases }
\label{fig6}
\end{figure}

Graph databases, such as Neo4j, NebulaGraph, and JanusGraph, utilize nodes and edges to store data, each employing their own unique storage mechanisms. They organize graph data within files, often in the form of arrays, which can be readily converted to a “\{node: attributes\}, \{edge: attributes\}” structure, as illustrated in \autoref{fig6}. This array-based storage approach is particularly well-suited to the retrieval techniques employed in our alignment method, preventing the need for repeated queries to the graph database during alignment and consequently reducing memory consumption.

\section{Evaluation Metrics Definition}
\label{app:metrics}


Given the complexity of graph databases, where multiple natural languages can describe a single GQL and vice versa, traditional NL2SQL evaluation metrics like Logical and Execution Accuracy are insufficient. GQL's intricate structure, capable of yielding diverse query results, and the variability in functional keywords for identical natural language queries necessitate a tailored evaluation approach. We address this by proposing three key questions, each leading to specific evaluation metrics:

\begin{itemize}
    \item {$\mathcal{Q}$1:} Evaluation of the syntax of generated GQLs.
    \item {$\mathcal{Q}$2:} Assessment of the model's semantic understanding.
    \item {$\mathcal{Q}$3:} Determination of query information accuracy.
\end{itemize}

\makeatletter
\newcommand{\removelatexerror}{\let\@latex@error\@gobble}
\makeatother
\begin{figure}[!t]
	\removelatexerror
        
	\begin{algorithm}[H]
		\caption{Combined Similarity}
            \label{alg1}
		\LinesNumbered
		\KwIn{$gold\_result, gql\_result, alpha, beta$ }
		\KwOut{Combined similarity $combinedSim$}
		$(tokens1, tokens2) \gets tokenize(gold\_result, gql\_result)$;
		$jaccardSim \gets \frac{|tokens1 \cap tokens2|}{|tokens1 |}$\;
		$tfidfVectors \gets computeTFIDF([sentence1, sentence2])$\;
		$bm25Sim \gets computeBM25(tfidfVectors)$\;
		$jaccardSim \gets jaccardSim / 1.0$\;
		$bm25Sim \gets (bm25Sim + 1) / 2.0$\;
		$bert\_score \gets cal\_bert\_score(gold\_result,gql\_result)$\;
		$combinedSim \gets beta * [(alpha * jaccardSim) + ((1-alpha) * bm25Sim) ]+ (1-beta)*bert\_score$\;
		\Return{$combinedSim$}\;
	\end{algorithm}
\end{figure}

For $\mathcal{Q}$1, we introduce the Syntax Accuracy (SA) metric, assessing if the generated GQL can be executed without syntax errors by the graph database:

\begin{equation}
SA=\frac{\text{Number of error-free GQLs}}{\text{Total number of test dataset}}
\end{equation}

To tackle $\mathcal{Q}$2, the Comprehension Accuracy (CA) metric measures the similarity between model-generated and gold standard GQLs, employing the text-embedding-ada-002 model for code similarity comparisons via cosine similarity.

For $\mathcal{Q}$3, we propose Execution Accuracy (EA) and Intra Execution Accuracy (IEA) metrics. EA evaluates global execution accuracy, while IEA assesses accuracy among syntactically correct GQLs. Considering GQL's diverse result formats, we adopt an enhanced Jaccard algorithm and BM25 for content completeness, and BertScore for semantic similarity, averaging the scores for a comprehensive evaluation. IEA, detailed in \autoref{alg1}, focuses on the accuracy of query results from correctly generated GQLs.

\section{Examples of Generation ERROR}
\label{app:error}
We have categorized the errors into three major categories and six minor categories. The major categories are: Ranker Error, Rewriter Error, and Refiner Error. These are further subdivided into Schema Selection Error, Skeleton Selection Error, No Related Information, Syntax Error, Query Misunderstanding, and Other. Specific details can be found in Table \ref{tab:errors}, and an analysis of the error statistics is provided in the main text of the paper.

\begin{landscape}
\begin{table}[h]
\centering
\caption{Error Types and Examples}
\begin{tabularx}{\linewidth}{@{} >{\hsize=.6\hsize}X >{\hsize=1\hsize}X >{\hsize=1.0\hsize}X >{\hsize=1.0\hsize}X >{\hsize=1.0\hsize}X @{}}
\toprule
\textbf{Error Type} & \textbf{Detail Type} & \textbf{Query} & \textbf{Gold GQL} & \textbf{Result GQL} \\ 
\midrule
\multirow{2}{=}[2.5ex]{Ranker Error} & Schema Selection Error & Who is Theseus Scamander's fiancee? & MATCH (n: character \{name: 'Theseus Scamander'\}) - \colorbox{lightblue}{[e: kindred {rel\_type: 'fiancee'}]} - (n1) return n1 & MATCH (v: character \{name: "Theseus Scamander"\}) -- (v2: character) WHERE \colorbox{lightred}{v2.marital=="fiancee"}  RETURN v2.name; \\ 
\cmidrule(l){2-5} 
 & Skeleton Selection Error & Find the first entity that Tim Duncan likes & GO FROM "Tim Duncan" OVER like \colorbox{lightblue}{LIMIT 1} & FETCH PROP ON player "Tim Duncan" -> \colorbox{lightred}{var1} return var1 \\ 
\midrule
Rewriter Error & No Related Information & May I ask if you can help me find all the entities that Porzingis likes and give me their IDs? & GO FROM \colorbox{lightblue}{"Kristaps Porzingis"} OVER like YIELD id(\$\$) AS vid | RETURN -.vid AS dst & GO FROM \colorbox{lightred}{"Porzingis"} OVER like YIELD dst(edge) AS id \\ 
\midrule
\multirow{3}{=}[3.75ex]{Refiner Error} & Syntax Error & May I ask if you can help me find players who are 29.5 years old or older? I need their ID and age information & LOOKUP ON player WHERE \colorbox{lightblue}{player.age} >= 29.5 YIELD id(vertex) as name, player.age AS Age & LOOKUP ON player WHERE \colorbox{lightred}{age} >= 29.5 YIELD id(vertex) as ID, player.age as Age \\ 
\cmidrule(l){2-5} 
 & Query Misunderstanding & Which department should I go to if I have hepatitis C virus infection and glomerulonephritis? & GO FROM \colorbox{lightblue}{"hepatitis C virus infection} \colorbox{lightblue}{and glomerulonephritis"} OVER cure\_department YIELD dst(edge) & MATCH (v1:disease\{name: \colorbox{lightred}{"hepatitis C virus infection"}\})-[:cure\_department]->(v2:department), (v3:disease \{name: \colorbox{lightred}{"glomerulonephritis"}\})-[:cure\_department]->(v4:department) RETURN v2.name, v4.name \\ 
\cmidrule(l){2-5} 
 & Other & Identify the entities that indirectly like Kobe Bryant communication, and then return the names of these entities & GO 2 STEPS FROM 'Kobe Bryant' OVER \colorbox{lightblue}{like REVERSELY} YIELD \$\$.player.name & GO 2 STEPS FROM "Kobe Bryant" OVER \colorbox{lightred}{like} YIELD \$\$.player.name AS Name \\ 
\bottomrule
\end{tabularx}
\label{tab:errors}
\end{table}

\end{landscape}

\end{document}